\documentclass[twoside,11pt]{article}

\usepackage{jmlr2e}
\usepackage{lastpage}
\usepackage{microtype}
\usepackage[english]{babel}
\usepackage{url}
\usepackage{amsmath}
\usepackage{booktabs}
\usepackage{cleveref}
\usepackage{tikz}
\usepackage{tkz-graph}

\usetikzlibrary{
    arrows,
    arrows.meta,
    fadings,
    patterns,
    positioning,
    shapes.geometric
}

\usetikzlibrary{arrows,topaths,calc}
\usetikzlibrary{positioning}
\usetikzlibrary{shadows}
\usetikzlibrary{shapes}
\usetikzlibrary{backgrounds}
\usetikzlibrary{decorations.pathreplacing}

\jmlrheading{23}{2022}{1-\pageref{LastPage}}{1/21; Revised 5/22}{9/22}{21-0000}{Author One and Author Two}

\jmlrheading{26}{2025}{1-\pageref{LastPage}}{7/24; Revised 1/25}{2/25}{24-1113}{Caglar Demir, Alkid Baci, N'Dah Jean Kouagou, Leonie Nora Sieger, Stefan Heindorf, Simon Bin, Lukas Blübaum, Alexander Bigerl, and Axel-Cyrille Ngonga Ngomo}

\ShortHeadings{Ontolearn}{Demir, Baci, Kouagou, Sieger, Heindorf, Bin, Blübaum, Bigerl, Ngonga Ngomo}

\begin{document}

\title{Ontolearn---A Framework for Large-scale\\ OWL Class Expression Learning in Python}

\author{\name Caglar Demir \email caglar.demir@upb.de
       \AND
       \name Alkid Baci \email alkid@campus.uni-paderborn.de
       \AND
       \name N'Dah Jean Kouagou \email ndah.jean.kouagou@upb.de
       \AND
        \name Leonie Nora Sieger \email leonie.sieger@upb.de
        \AND
        \name Stefan Heindorf \email heindorf@upb.de
       \AND
       \name Simon Bin \email sbin@informatik.uni-leipzig.de
       \AND
       \name Lukas Blübaum \email lukasbl@campus.uni-paderborn.de
       \AND
       \name Alexander Bigerl \email alexander.bigerl@upb.de
       \AND
       \name Axel-Cyrille Ngonga Ngomo \email axel.ngonga@upb.de \\
       \addr Department of Computer Science \\
       Paderborn University\\
       Warburger Str. 100, 33098 Paderborn, Germany}

\editor{Sebastian Schelter}

\maketitle

\begin{abstract}
In this paper, we present Ontolearn---a framework for learning OWL class expressions over large knowledge graphs.
Ontolearn contains efficient implementations of recent state-of-the-art symbolic and neuro-symbolic class expression learners including EvoLearner and DRILL.
A learned OWL class expression can be used to classify instances in the knowledge graph.
Furthermore, Ontolearn integrates a verbalization module based on an LLM to translate complex OWL class expressions into natural language sentences.
By mapping OWL class expressions into respective SPARQL queries, Ontolearn can be easily used to operate over a remote triplestore.
The source code of Ontolearn is available at \url{https://github.com/dice-group/Ontolearn}.

\end{abstract}


\section{Introduction}

Explainability is quintessential to establishing trust in AI decisions~\citep{Rudin2019Stop}. 
It becomes particularly important when an AI algorithm relies on data from the Web---the largest and arguably most used information infrastructure available to humanity with over 5 billion users~\citep{Demir2023DRILL}.
A key development over the last decade has been the increasing availability of Web data in the form of large-scale knowledge bases in RDF~\citep{Hogan2022Knowledge}.
Although devising explainable ML approaches for Web-scale RDF knowledge graphs (KGs) is an indisputable building block of a trustworthy Web, most symbolic learners cannot operate well on large KGs having millions of triples.

Given a knowledge base and sets of positive and negative examples, the goal of class expression learning is to learn a class expression in description logics such that the positive examples are instances of this expression and the negative examples are not (see Figure~\ref{fig:family_kb} for an example and \citet{Lehmann2010Concept} for a formal definition).
DL-Learner~\citep{Lehmann2009DL-Learner} was regarded as the most mature system for OWL class expression learning~\citep{Buhmann2018DL-Learner,Sarker2019Efficient}, featuring the symbolic learners ELTL, OCEL, and CELOE.
However, because DL-Learner has not been actively maintained since its last release in 2021, it does not integrate the latest neuro-symbolic models.

\begin{figure*}
    \centering
    \small
    \includegraphics[width=\textwidth]{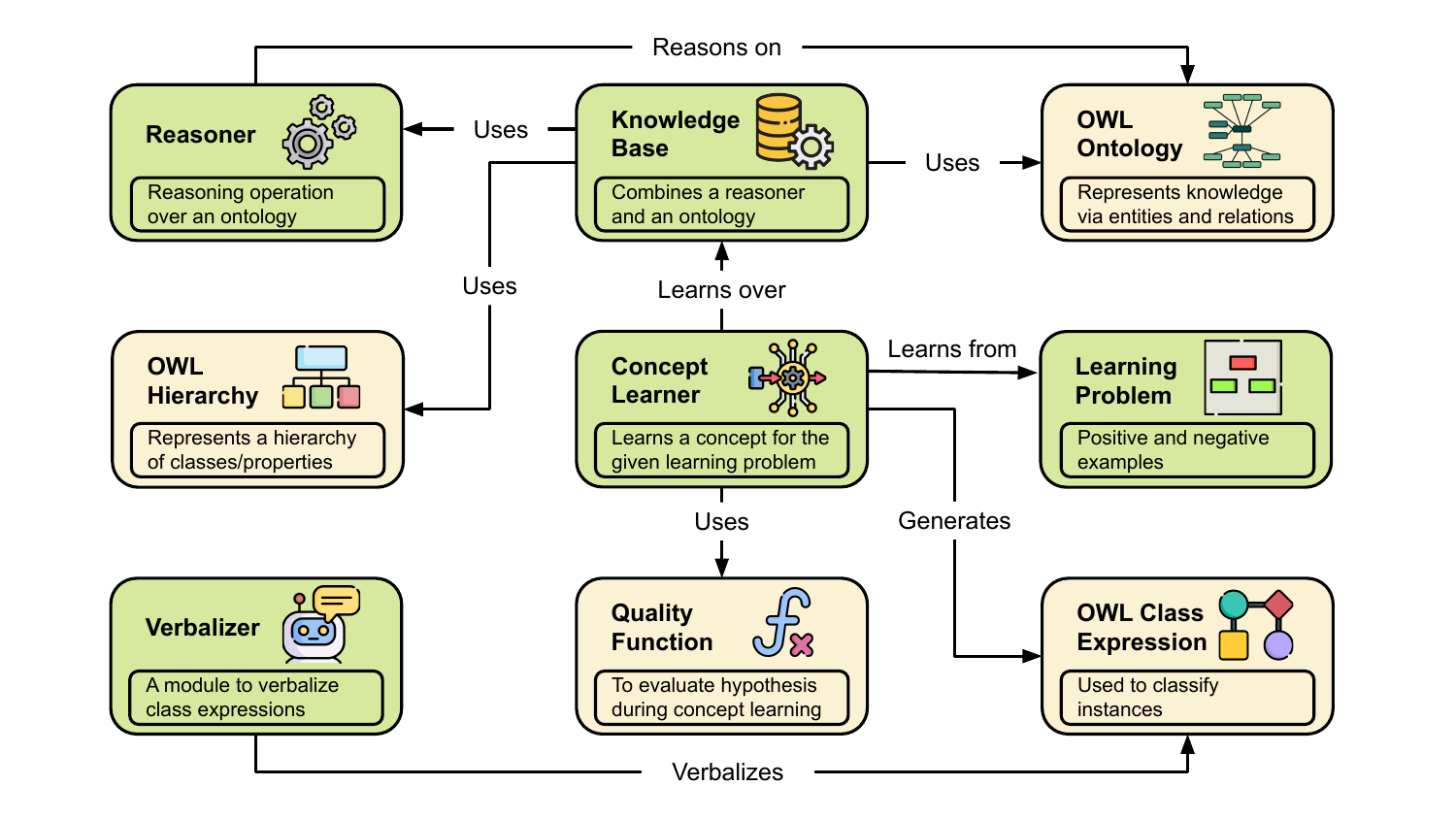}
    \caption{Ontolearn architecture. Green rectangles show the top-level components, whereas beige rectangles show the sub-components.} 
    \label{OntolearnComponents}
\end{figure*}

In this paper, we present Ontolearn, an open-source Python library that facilitates OWL class expression learning over large RDF knowledge graphs. 
\Cref{OntolearnComponents} shows the software architecture of Ontolearn.
Ontolearn
\begin{enumerate}
    \item provides nine recent state-of-the-art symbolic, neuro-symbolic and deep learning algorithms to learn OWL class expressions along with efficient Python implementations of CELOE and OCEL from DL-Learner,
    \item uses various OWL reasoners for class expression learning via Owlapy,\footnote{\url{https://pypi.org/project/owlapy}}
    \item verbalizes complex OWL class expressions through large-language models (LLMs).
\end{enumerate}

At the time of writing, Ontolearn provides more OWL class expression learning algorithms than any other publicly available framework.
Moreover, Ontolearn is a well-tested framework that comes with 156 unit and regression tests along with 95\% test coverage.
Ontolearn has already been downloaded more than 26,000 times.
Importantly, we provide 26 example scripts along with a documentation to guide new users.\footnote{\url{https://ontolearn-docs-dice-group.netlify.app/}}
Ontolearn can easily be used via PyPI under the MIT license.%
\footnote{\url{https://pypi.org/project/ontolearn}}

\section{Ontolearn}

The \textbf{Knowledge Base} class is designed for efficient and easy access to a given OWL ontology. 
We support reading OWL ontologies into memory in RDF/XML, OWL/XML, and N-Triples formats.
Given the challenges of loading large-scale ontologies with more than $10^8$ triples into memory, data can alternatively be loaded into a triplestore like Tentris \citep{Bigerl2020Tentris} and OWL class expressions can be mapped to SPARQL queries \citep{Karalis2024Evaluating}, enabling efficient instance retrieval from the triplestore. 
Users can either provide the file path of an ontology or the endpoint of a triplestore when using Ontolearn.

\textbf{Reasoner} is an external component used to infer new knowledge from asserted axioms and to determine all instances of a given class expression. 
During concept learning in Ontolearn, we use OWL reasoners, e.g. Hermit~\citep{Glimm2014HermiT} and Pellet~\citep{Sirin2007Pellet}), which are implemented in Owlapy, to retrieve instances of class expressions.

\begin{figure}
    \centering
    \small
    \definecolor{babyblue}{RGB}{53,172,229}
\definecolor{turquoise}{RGB}{0, 245, 255}

\tikzset{
    man/.pic={    
        \node[babyblue, circle, fill, minimum size = 1.5mm, inner sep = 0] (head) at (0,0) {};
        \node[
            babyblue,
            draw,
            fill,
            rectangle,
            rounded corners = 0.2mm,
            minimum width = 1.2mm,
            minimum height = 2.4mm,
            below = 0.1mm of head,
            inner sep = 1pt
        ] (body) {};
        \draw[babyblue, line width = 0.5mm,  round cap-round cap]
            ([shift={(-0.42mm, 0.45mm)}]body.south) --++(-90:2.4mm);
        \draw[babyblue, line width = 0.5mm,round cap-round cap]
            ([shift={(0.42mm, 0.45mm)}]body.south) --++(-90:2.4mm);
        \draw[babyblue, line width = 0.2mm, round cap-round cap, rounded corners]
            ([shift = {(-0.85mm, -0.2mm)}]body.north) -- ++(0mm, -2mm);
        \draw[babyblue, line width = 0.2mm, round cap-round cap, rounded corners]
            ([shift = {(0.85mm, -0.2mm)}]body.north) -- ++(0mm, -2mm);
    }
}
\tikzset{
    woman/.pic={    
        \node[magenta, circle, fill, minimum size = 1.5mm, inner sep = 0] (head) at (0,0) {};
        \node[
            magenta,
            transform shape,
            draw,
            fill,
            trapezium,
            trapezium angle=65,
            trapezium stretches=true,
            rounded corners = 0.2mm,
            minimum width = 2mm,
            minimum height = 2.8mm,
            below = 0.1mm of head,
            inner sep = 1pt
        ] (body) {};
        \draw[magenta, line width = 0.5mm,  round cap-round cap]
            ([shift={(-0.35mm, 0.45mm)}]body.south) --++(-90:2mm);
        \draw[magenta, line width = 0.5mm,round cap-round cap]
            ([shift={(0.35mm, 0.45mm)}]body.south) --++(-90:2mm);
        \draw[magenta, line width = 0.2mm, round cap-round cap, rounded corners]
            ([shift = {(-0.8mm, -0.2mm)}]body.north) -- ++(-0.35mm, -2mm);
        \draw[magenta, line width = 0.2mm, round cap-round cap, rounded corners]
            ([shift = {(0.8mm, -0.2mm)}]body.north) -- ++(0.35mm, -2mm);
    }
}

\newcommand{\Male}[4]{%
    \node[label = {[font = \sf\tiny, label distance = -1mm]below:#4}] (#1) at (#2, #3) {%
        \begin{tikzpicture}
            \pic at (0, 0) {man};
        \end{tikzpicture}
    };
}

\newcommand{\Female}[4]{%
    \node[label = {[font = \sf\tiny, label distance = -1mm]below:#4}] (#1) at (#2, #3) {%
        \begin{tikzpicture}
            \pic at (0, 0) {woman};
        \end{tikzpicture}
    };
}

\newcommand{\cellwidth}{0.32\textwidth}

        \tikzset{BaseCell/.style = {
                draw = white,
                inner sep = 5pt,
                font = \tiny\sf,
                align = left,
                line width = 1pt,
                rounded corners = 3pt
            }}
        \tikzset{
            Example/.style = {
                BaseCell,
                draw = white,
                fill = black!70,
                text = white
            }
        }%
        \tikzset{
            Graph/.style = {
                fill = black!20
            }
        }%
        \tikzset{
            KB/.style = {
                BaseCell,
                fill = white
            }
        }%

     \begin{tikzpicture}[
                    xscale = 1.4,
                    yscale = 1.7,
                    TreeLine/.style = {
                        line width = 0.4mm,
                        black!70
                    },
                    HighlightLine/.style = {
                        TreeLine,
                        orange
                    },
                    InferredLine/.style = {
                        HighlightLine,
                        dashed
                    },
                    Marriage/.style = {
                        circle,
                        fill = black!70,
                        inner sep = 0pt,
                        minimum width = 3
                    }
                ]
                
                    \Male  {idF10M171}{1}{3}{F10M171}
                    \Female{idF10F172}{2}{3}{F10F172}
                    
                    \Male  {idF10M180}{0}{2}{F10M180}
                    \Female{idF10F179}{1}{2}{F10F179}
                    \Male  {idF10M173}{2}{2}{F10M173}
                    \Female{idF10F174}{3}{2}{F10F174}

                    \Female{idF10F177}{2}{1}{F10F177}
                    \Female{idF10F175}{3}{1}{F10F175}

                    \draw[TreeLine] (idF10M171) -- (idF10F172);
                    \draw[TreeLine] (idF10M180)   -- (idF10F179);
                    \draw[TreeLine] (idF10M173)  -- (idF10F174);    
        
                    \node[Marriage] at (1.5, 3) {};            
                    \draw[TreeLine] (1.5, 2.6) -- (1.5, 3);    
                    \draw[TreeLine] (idF10F179)                   
                                    -- (1, 2.4)
                                    arc (180:90:0.1)
                                    -- (1.4, 2.5)
                                    arc (270:360:0.1)
                                    arc (180:270:0.1)
                                    -- (1.9, 2.5)
                                    arc (90:0:0.1)
                                    -- (idF10M173);

                    \node[Marriage] at (0.5, 2) {};            
                    
                    \node[Marriage] at (2.5, 2) {};            
                    \draw[TreeLine] (2.5, 1.6) -- (2.5, 2);    
                    \draw[TreeLine] (idF10F177)                  
                                    -- (2, 1.4)
                                    arc (180:90:0.1)
                                    -- (2.4, 1.5)
                                    arc (270:360:0.1)
                                    arc (180:270:0.1)
                                    -- (2.9, 1.5)
                                    arc (90:0:0.1)
                                    -- (idF10F175);

                    \node[Marriage, black] at (1.5, 3) {};
                    \node[Marriage, black] at (2.5, 2) {}; 
                    \draw[TreeLine] (idF10F179)
                                         -- (1, 2.4)
                                         arc (180:90:0.1)
                                         -- (1.4, 2.5)
                                         arc (270:360:0.1)
                                         -- (1.5, 3)
                                         -- (idF10F172)
                                         -- (1.5, 3)
                                         -- (1.5, 2.6)
                                         arc (180:270:0.1)
                                         -- (1.9, 2.5)
                                         arc (90:0:0.1)
                                         -- (idF10M173)
                                         -- (2.5, 2)
                                         -- (2.5, 1.6)
                                         arc (180:270:0.1)
                                         -- (2.9, 1.5)
                                         arc (90:0:0.1)
                                         -- (idF10F175);
                     \node[KB] (kb1) {%
                       \vspace{-4cm}
                       \hspace{4cm} \begin{minipage}{\cellwidth}
                            \textbf{Learning Problem - ``Married Female'':} \\
                            $E^{+} = \{F10F172, F10F179, F10F174\}$  \\
                            $E^{-} = \{F10F177, F10F175\}$ \\
                            \textbf{Learned Concept:} \\
                            Learned concept in DL: $\text{\textit{Female}} \sqcap (\exists \text{\textit{married.}}\top)$ \\
                            LLM-verbalized: \textit{A female who is married} \\
                        \end{minipage}
                    };
                \end{tikzpicture}
        \begin{tikzpicture}
        [Line/.style = {inner sep = 2pt,line width = 0pt}]        

            \node[Line] (line1) {%
                \begin{tikzpicture}
                    \node[KB] (kb1) {%
                        \begin{minipage}{\cellwidth}
                            \textbf{TBox:} \\
                            Brother $\sqsubseteq$Male\\
                            Brother $\sqsubseteq$ 
                            PersonWithASibling\\
                            Child $\sqsubseteq$ Person\\
                            Daughter $\sqsubseteq$ Child, Daughter $\sqsubseteq$ Female\\
                            Father $\sqsubseteq$ Male, Father $\sqsubseteq$ Parent\\
                            Female $\sqsubseteq$ Person\\
                            Grandchild $\sqsubseteq$ Child\\
                            Granddaughter $\sqsubseteq$ Female\\ Granddaughter $\sqsubseteq$ Grandchild\\
                            Grandfather$\sqsubseteq$ Grandparent\\ Grandfather$\sqsubseteq$ Male\\
                            Grandmother$\sqsubseteq$Female\\ 
                            Grandmother$\sqsubseteq$ Grandparent\\
                            Grandparent$\sqsubseteq$Parent\\
                            Grandson$\sqsubseteq$Grandchild,Grandson$\sqsubseteq$Male\\
                            Male $\sqsubseteq$ Person\\
                            Mother $\sqsubseteq$ Person, Mother $\sqsubseteq$ Parent\\
                            Parent $\sqsubseteq$ Person\\                            PersonWithASibling $\sqsubseteq$ Person\\
                            Sister $\sqsubseteq$ Female\\
                            Sister $\sqsubseteq$ PersonWithASibling\\
                            Son $\sqsubseteq$ Child, Son $\sqsubseteq$ Male
                            
                        \end{minipage}
                    };
                \end{tikzpicture}
            };
        \end{tikzpicture}
       
    \caption{A partial visualization of the Family knowledge base along with a learning problem defined by $E^+$ and $E^-$. An example of a learned concept is given in DL syntax which is verbalized into natural language using an LLM.}
     \label{fig:family_kb}
\end{figure}
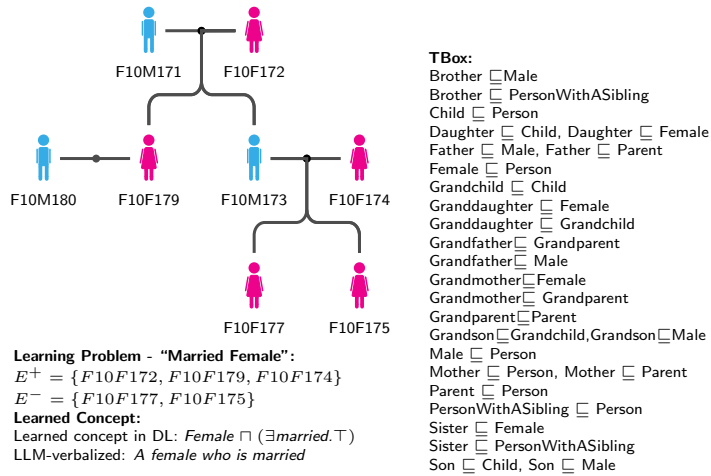

The \textbf{Learning Problem} class encapsulates all relevant information related to OWL class expression learning problems.
\Cref{fig:family_kb} visualizes a supervised OWL class expression learning problem based on positive $E^+$ and negative $E^-$ examples. 
Ontolearn provides scalable implementations of recent state-of-the-art OWL class expression learners, including EvoLearner~\citep{Heindorf2022EvoLearner}, CLIP~\citep{Kouagou2022Learning}, NCES~\citep{Kouagou2023NCES}, NCES2~\citep{kouagou2023nces2}, DRILL~\citep{Demir2023DRILL}, Nero~\citep{demir2023learning}, ROCES~\citep{kouagou2024roces} and efficient Python implementations of CELOE and OCEL~\citep{Lehmann2009DL-Learner}.
Moreover, Ontolearn includes wrappers for the DL-Learner framework, allowing access to the original implementations of the previous state-of-the-art symbolic learners CELOE and OCEL.
Ontolearn also supports sampling techniques to accelerate the learning process~\citep{Baci2023Accelerating}, and triplestore implementations for loading and querying large-scale knowledge bases in distributed and cloud-based environments.
Finally, Ontolearn offers the functionality of verbalizing class expressions into natural language using the power of LLMs like Llama~\citep{touvron2023llama} or Mistral~\citep{jiang2023mistral}.

\section{Implementation}
The documentation%
\footnote{\url{https://ontolearn-docs-dice-group.netlify.app/}}
gives an overview of Ontolearn's key functions and how to use them.
The Ontolearn project consists of approximately 20,000 lines of code.
At the time of writing, it contains 26 example scripts showcasing its main functionalities. 
The code's correctness is ensured by 156~test cases in Python's \emph{unittest} framework. 

Ontolearn provides all the functionalities and helper functions to easily extend it with novel concept learners.
It also includes the code required to launch it as a web service.

\section{Use Cases \& Complementary Libraries}
Ontolearn has already been applied in industrial projects, where ante-hoc explainability is required.
For example, Ontolearn has been used within Industry 4.0 settings~\citep{demir2022rapid} to automatically learn human-interpretable descriptions of the skills of machines, which is crucial for tasks like skill matching in production processes.

EDGE~\citep{Sapkota2024EDGE} employs Ontolearn to explain graph neural networks in terms of class expressions.
AutoCL~\citep{Li2024AutoCL} facilitates feature selection and hyperparameter tuning for Onotlearn's concept learners.
OntoSample~\citep{Baci2023Accelerating} applies advanced graph sampling techniques prior to inputting the ontology into Ontolearn, enabling substantial speedups while maintaining high predictive performance.
Tab2Onto~\citep{zahera2022tab2onto} allows to automatically convert tabular data to an OWL ontology, simplifying the application of Ontolearn to a wide range of industrial use cases. 

\section*{Acknowledgment}
This work has received funding from the European Union's Horizon 2020 research and innovation programme within the project KnowGraphs under the Marie Skłodowska-Curie grant No 860801,
the European Union’s Horizon Europe research and innovation programme within the project ENEXA under the grant No 101070305,
the European Union’s Horizon Europe research and innovation programme within the project LEMUR under the Marie Skłodowska-Curie grant agreement No 101073307,
and the Lamarr Fellow Network funded project WHALE (LFN 1-04).
We acknowledge the contributions of the following creators for their icons used in \Cref{OntolearnComponents}: freepik, becris, gravisio, and alla-afanasenko, available at \url{https://www.flaticon.com}.

\end{document}